\documentclass[runningheads]{llncs}
\usepackage{graphicx}
\usepackage{comment}
\usepackage{amsmath,amssymb} 
\usepackage{color}
\usepackage{multirow}
\usepackage{hyphenat}
\usepackage{hyperref}
\DeclareMathOperator*{\argmin}{arg\,min}

\newcommand{\eat}[1]{}

\usepackage{pifont}
\newcommand{\cmark}{\ding{51}}%
\newcommand{\xmark}{\ding{55}}%

\begin{document}
\pagestyle{headings}
\mainmatter
\def\ECCVSubNumber{6837}  

\title{Improving Object Detection with\\ \emph{Selective} Self-Supervised Self-Training} 

\titlerunning{Improving Object Detection with Selective Self-supervised Self-training}
%
\authorrunning{Yandong Li, Di Huang, Danfeng Qin, Liqiang Wang, Boqing Gong}


\author{Yandong Li\inst{1,2,*}  \and Di Huang\inst{2} \and Danfeng Qin\inst{2} \and  Liqiang Wang\inst{1} \and Boqing Gong\inst{2}
}
\institute{$^1$University of Central Florida, $^2$Google Inc
\\
\email{ lyndon.leeseu@outlook.com, lwang@cs.ucf.edu, \{dihuang,qind,bgong\}@google.com}
}

\maketitle

\begin{abstract}
We study how to leverage Web images to augment human-curated object detection datasets. Our approach is two-pronged. On the one hand, we retrieve Web images by image-to-image search, which incurs less domain shift from the curated data than other search methods. The Web images are diverse, supplying a wide variety of object poses, appearances, their interactions with the context, etc. On the other hand, we propose a novel learning method motivated by two parallel lines of work that explore unlabeled data for image classification: self-training and self-supervised learning. They fail to improve object detectors in their vanilla forms due to the domain gap between the Web images and curated datasets. To tackle this challenge, we propose a selective net to rectify the supervision signals in Web images. It not only identifies positive bounding boxes but also creates a safe zone for mining hard negative boxes. We report state-of-the-art results on detecting backpacks and chairs from everyday scenes, along with other challenging object classes. 
{\let\thefootnote\relax\footnote{{$^{*}$ This work was done while the first author was an intern at Google Inc.}}}
\end{abstract}
\section{Introduction}
\vspace{-6pt}
%

Object detection is a fundamental task in computer vision. It has achieved unprecedented performance for many objects, partially thanks to the recently developed deep neural detectors~\cite{ren2015faster,lin2017feature,redmon2017yolo9000,tan2020efficientdet}. Some  detectors have made their way into real-world applications, such as smart mobile phones  and self-driving cars. 

However, upon a careful investigation into the top three teams' class-wise detection results on  80 common objects in context (COCO)~\cite{lin2014microsoft}, we find that they still fall short of detecting backpacks, handbags, and chairs, among other \emph{functional} objects. As of the paper submission, they report an average precision~\cite{lin2014microsoft} of less than 0.30 on detecting backpacks and less than 0.40 on chairs. 

\begin{figure*}
  \centering
  \includegraphics[width = 0.98\textwidth]{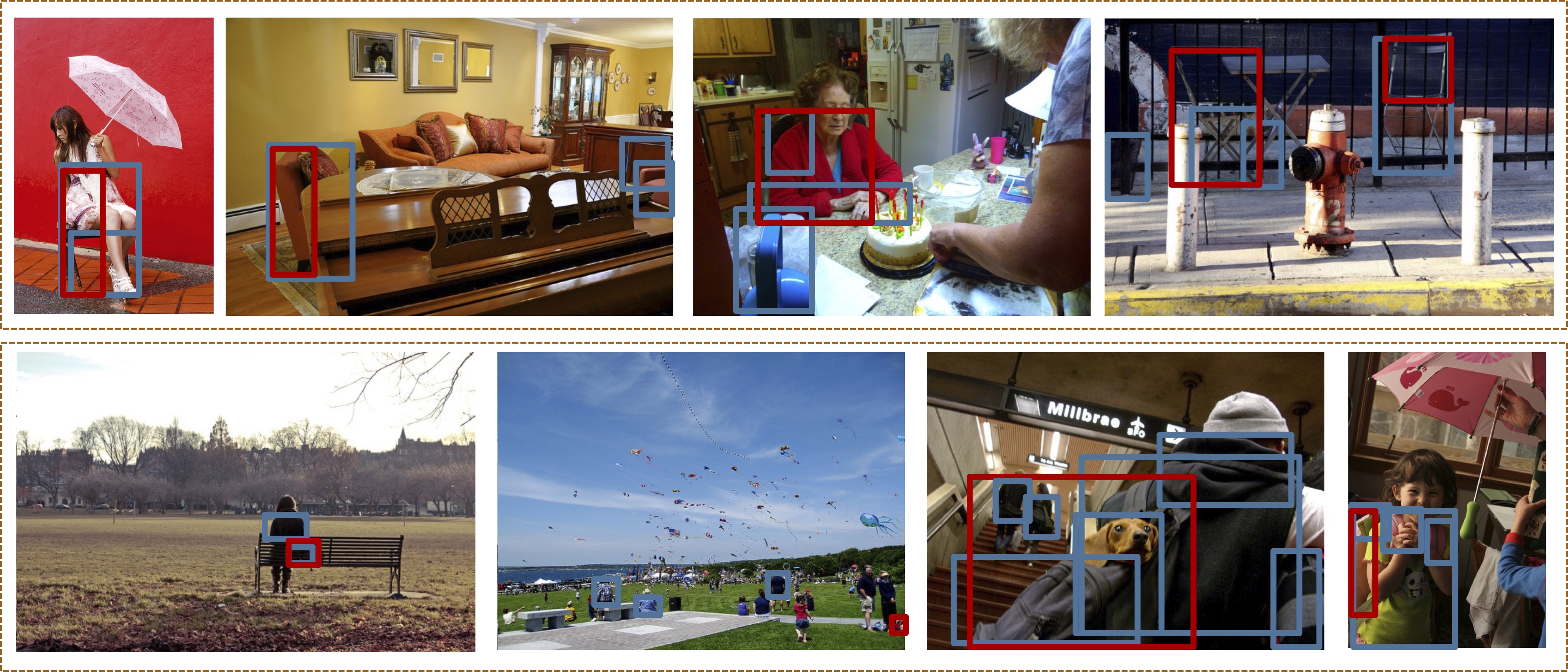}
  \vspace{-10pt}
  \caption{Chairs (top) and backpacks (bottom) are difficult to detect in daily scenes. (Images and groundtruth boxes (red) are from COCO~\cite{lin2014microsoft}, and the blue boxes are predicted by the best detectors in our experiments.)}
  \label{fig:badpre}
  \vspace{-15pt}
\end{figure*}

These man-made objects are defined by their functionalities more than visual appearances, leading to high intra-class variation.  Minsky~\cite{minsky1988society} wrote that {``there's little we can find in common to all chairs – except for their intended use.''}  Grabner \textit{et al.}~\cite{grabner2011makes} quantitatively evaluated the challenges of detecting functional objects like chairs. Another potential reason that contributes to the low performance in detecting backpacks and chairs is that they are too common to draw photographers' attention. Consequently, they often sit out of the camera focus, appear small, and become occluded in context (cf.\  Figure~\ref{fig:badpre}).

\emph{How to improve the detection of backpacks, chairs, and other common, ``less eye-catching'' functional objects?} We believe the answer resides on both the quality of training data and the inductive bias of advanced detectors. In this paper, we focus on the data aspect and mainly study the potential of  \emph{unlabeled} Web images for improving the detection results of backpacks and chairs, without heavily taxing human raters.

In other words, we study how to leverage the \emph{unlabeled} Web images to augment human-curated object detection datasets.  Web images are diverse and massive, supplying a wide variety of object poses, appearances, their interactions with the context, etc., which may lack in the curated object detection datasets. However, the Web images are \emph{out of the distribution} of the curated datasets. The domain gap between them calls for a careful design of methods to effectively take advantage of the signals in the Web data.

Our approach is two-pronged. On the one hand, we retrieve a big pool of candidate Web images via Google Image (\url{https://images.google.com}) by using its image-to-image search. The query set consists of all training images in the original human-curated dataset. Compared with text-based search, which mainly returns iconic photos, the image-based search gives rise to more natural images with diverse scenes, schematically reducing the domain mismatch between the retrieved images and the original datasets.

On the other hand, we propose a novel learning method to utilize the Web images for object detection by drawing inspiration from
self-training~\cite{scudder1965probability,xie2019self} and self-supervised learning~\cite{oord2018representation,donahue2019large,he2019momentum,wu2018unsupervised,chen2020simple,gidaris2018unsupervised}, both of which are popular in semi-supervised learning. Our problem is similar to semi-supervised learning, but there exists a domain gap between the Web images and the curated datasets. We find that the domain gap fails both self-training and  self-supervised learning in their vanilla forms because the out-of-domain Web images give rise to many inaccurate candidate boxes and uncalibrated box classification scores. To tackle these challenges, we propose a selective net to identify high-quality positive boxes and a safe zone for mining hard negative boxes~\cite{lin2017focal,ren2015faster}. It rectifies the supervision signals in Web images, enabling self-training and self-supervised learning to improve neural object detectors by leveraging the Web images. 


The main contributions in the paper are as follows. First, we customize self-training for the object detection task by a selective net, which identifies positive bounding boxes and assigns some negative boxes to a safe zone to avoid messing up the hard negative mining in the training of object detectors. Second, we improve the consistency-based~\cite{laine2016temporal,wei2018improving,ding2018semi} semi-supervised object detection~\cite{jeong2019consistency} by the selective net under our self-training framework. Third, to the best of our knowledge, this work is the first to explore \emph{unlabeled, out-of-domain}
Web images to augment curated object detection datasets. We report state-of-the-art results for detecting backpacks and chairs, along with other challenging objects.

\section{Augmenting COCO detection with Web images}
We augment the training set of COCO detection~\cite{lin2014microsoft} by retrieving relevant Web images through Google Image (\url{https://images.google.com}). We focus on the backpack and chair classes in this paper. They represent non-rigid and rigid man-made objects, respectively, and the existing results of detecting them are still unsatisfactory (less than 0.40 $AP$ on COCO as of March 5th, 2020). 

\vspace{-5pt}
\subsubsection{COCO-backpack, COCO-chair, Web-backpack, and Web-chair.}
\begin{figure*}
  \centering
  \includegraphics[width = 0.98\textwidth]{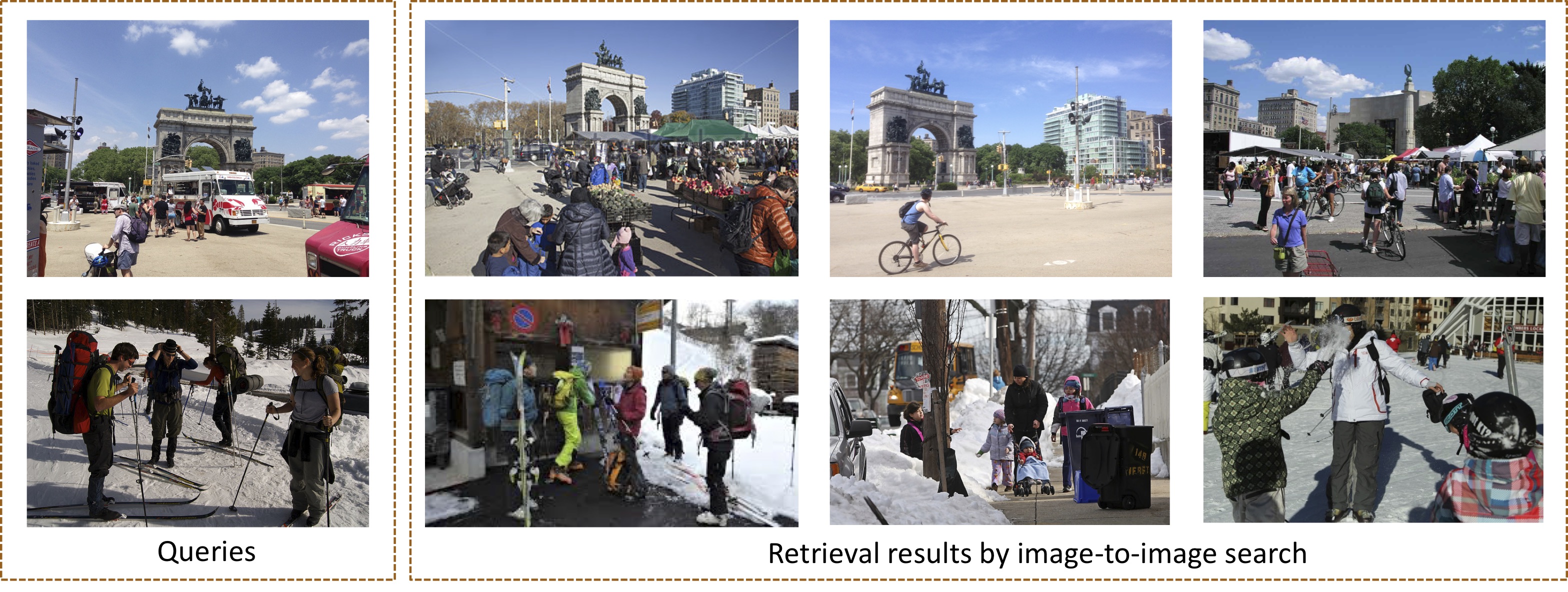}
  \vspace{-10pt}
  \caption{Examples of the top three retrieved images from image-to-image search.}
  \label{fig:example_query}
  \vspace{-15pt}
\end{figure*}

COCO is a widely used dataset for object detection, which contains 118k training images and 5k validation images~\cite{lin2014microsoft}. Out of them, there are 8,714 backpacks in 5,528 training images. We name these images the COCO-backpack query set. Similarly, we have a COCO-chair query set that contains 12,774 images with 38,073 chair instances. Using the images in COCO-backpack and COCO-chair to query Google Image, we obtain 70,438 and 186,192 unlabeled Web images named Web-backpack and Web-chair,  respectively. We have removed the Web images that are nearly duplicate with any image in the COCO training and validation sets.  Figure~\ref{fig:example_query} shows two query images and the retrieved Web images. 

\vspace{-5pt}
\subsubsection{Labeling Web-backpack\inst{1}.} To facilitate the evaluation of our approach and future research, we label a subset of the Web-backpack images. This subset contains 16,128 images and is selected as follows. We apply a pre-trained R101-FPN object detector~\cite{wu2019detectron2,lin2017feature} to all Web-backpack images and then keep the ones that contain at least one backpack box detected with the confidence score higher than 0.7. We ask three raters to label each survived images. One rater draws bounding boxes over all backpacks in an image. The other two examine the results sequentially. They modify the boxes if they find any problem with the previous rater's annotation. Please see the supplementary materials for more details of the annotation procedure, including a full annotation instruction we provided to the raters.


\begin{figure*}
\vspace{-10pt}
  \centering
  \includegraphics[width = 0.98\textwidth]{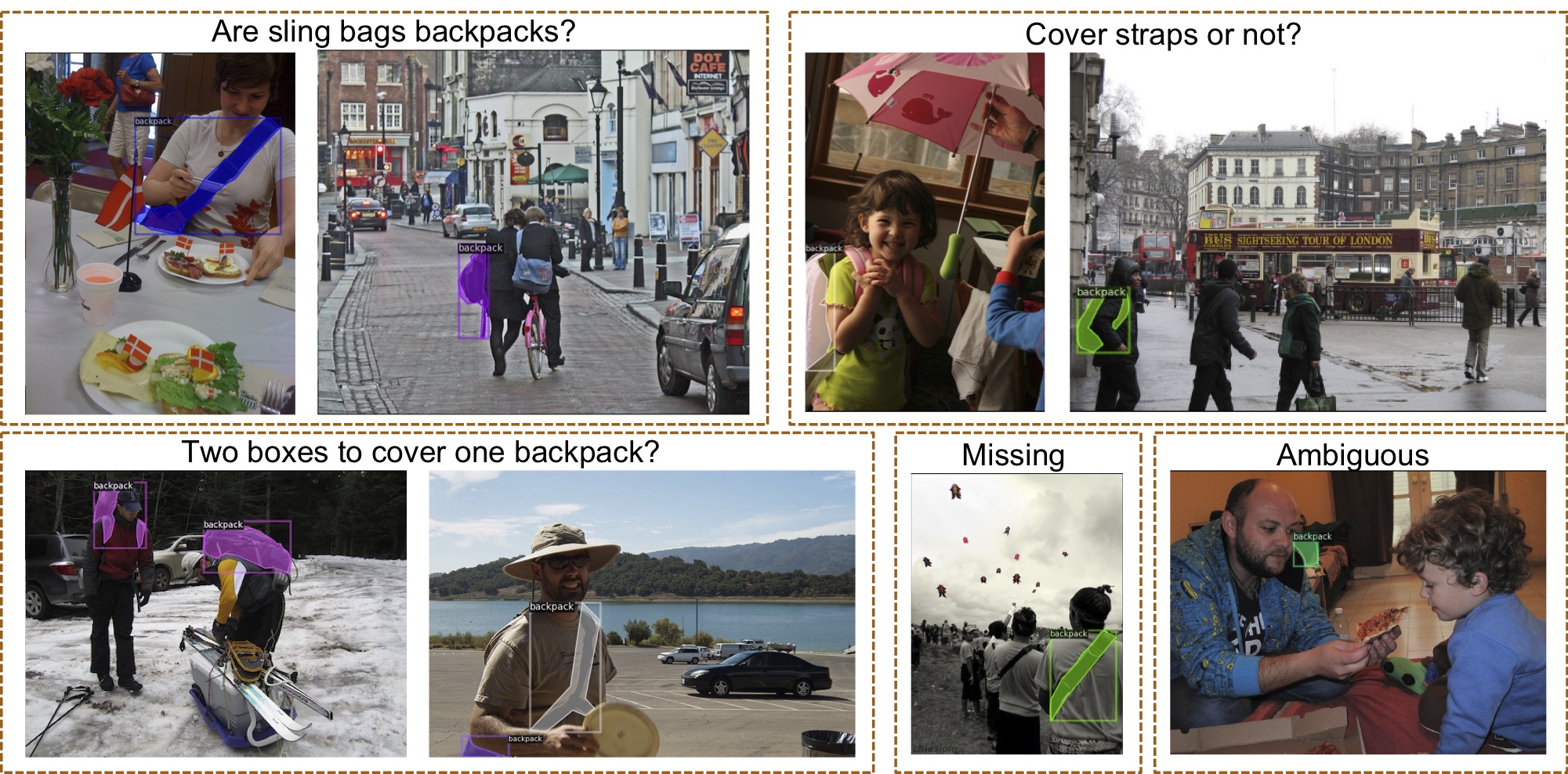}
  \vspace{-10pt}
  \caption{Labeling errors in COCO-backpack. Two images in one group mean that they contain conflict annotations.}
  \label{fig:coco_label}
  \vspace{-15pt}
\end{figure*}

\vspace{-5pt}
\subsubsection{Relabeling COCO-backpack\inst{1}.} Using the same annotation procedure above, we also relabel the backpacks in COCO training and validation sets. The main reason for the relabeling is to mitigate the annotation mismatch between Web-backpack and COCO-backpack caused by different annotation protocols. Another reason is that we observe inconsistent bounding boxes
in the existing COCO detection dataset. As Figure~\ref{fig:coco_label} shows, some raters label the sling bags, while others do not, and some raters enclose the straps in the bounding boxes and others do not. We still ask three raters to label the COCO-backpack training images. For the validation set, we tighten the quality control and ask five raters to screen each image. 

\begin{table*}
\centering
\caption{Statistics of the Web images in this paper and their counterparts in COCO}
\label{tab:stat}
\vspace{-10pt}
\resizebox{\textwidth}{!}{
\begin{tabular}{l|c|c|c|c}
\hline
 Dataset & $\#$ Images & $\#$ Boxes by COCO & $\#$ Boxes by us & $\#$ Raters\\
\hline
COCO-backpack& 5,528 & 8,714 & 7,170 & 3\\
COCO-backpack validation& 5,000 & 371 & 436 & 5\\
Web-backpack& 70,438 & -- & -- & --\\
Web-backpack labeled& 16,128 & -- & 23,683 & 3\\
\hline
COCO-chair& 12,774 & 38,073 & -- &--\\
COCO-chair validation& 5,000 & 1,771 & -- & --\\
Web-chair & 186,192 & -- & -- & --\\
\hline
\end{tabular}
}
\vspace{-10pt}
\end{table*}

{\let\thefootnote\relax\footnote{{$^{1}$ We use \href{http://cloud.google.com/ai-platform/data-labeling/docs}{\textcolor{blue}{Google Cloud Data Labeling Service}} for all the labeling work.}}}

\subsubsection{Statistics.} Table~\ref{tab:stat} shows the statistics of the datasets used in this paper. We augment the COCO-backpack (chair) by Web-backpack (chair), whose size is about 15 times as the former. ``Web-backpack labeled'' is for evaluation  only.

\section{\emph{Selective} self-supervised self-training}
\begin{figure*}
\vspace{-10pt}
  \centering
  \includegraphics[width = 0.84\textwidth]{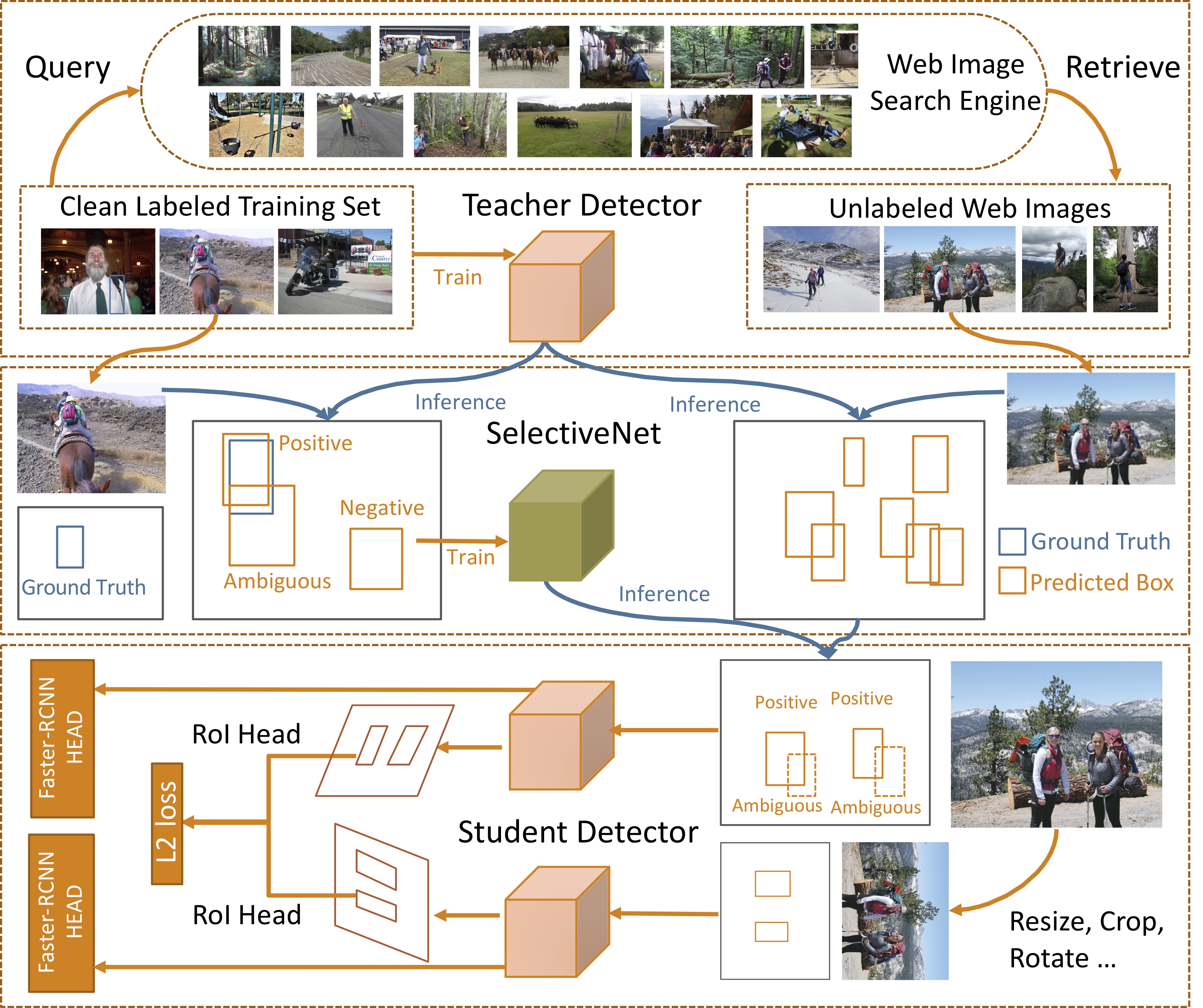}
  \vspace{-12pt}
  \caption{Overview of the proposed approach. Top: using a small curated dataset to retrieve relevant Web images and to train a teacher detector. Middle: training a selective net to group pseudo boxes predicted by the teacher into positive, negative, and ambiguous groups. Bottom: learning a student detector from the Web-augmented training set with a self-supervised loss.}
  \label{fig:overview}
  \vspace{-15pt}
\end{figure*}

In this section, we describe our learning method named Selective Self-Supervised Self-training for Object Detection (S$^4$OD). Without loss of generality, we consider detecting only one class of objects from an input image. Denote by $\mathcal{D} = \{(\mathbf{I}_1,\mathcal{T}_1),(\mathbf{I}_2,\mathcal{T}_2),...,(\mathbf{I}_n,\mathcal{T}_n)\}$ the labeled image set and $\mathcal{U} = \{\widetilde{\mathbf{I}}_1,\widetilde{\mathbf{I}}_2,...,\widetilde{\mathbf{I}}_m\}$ the crawled Web images, where $t_i^j = \{x_i^j,y_i^j,w_i^j,h_i^j\} \in \mathcal{T}_i$ contains the top-left  coordinate, width, and height of the $j$-th ground-truth bounding box in the $i$-th image $\mathbf{I}_i$. The labeled image set $\mathcal{D}$ is significantly smaller than the set of Web images $\mathcal{U}$. Besides, there exists a  domain shift between the two sets, although we have tried to mitigate the mismatch by using the image-to-image search. Finally, some Web images could contain zero objects of the class being considered. In the following, we first customize vanilla self-training for object detection, discuss its limitations and fixations by a selective net, and then arrive at the full S$^4$OD algorithm. Figure~\ref{fig:overview} illustrates a diagram of our approach. 

\subsection{Self-training for  object detection (SOD)} \label{sec:sod}

Given the labeled set  $\mathcal{D}$ and unlabeled set $\mathcal{U}$, it is natural to test how self-training performs, especially given that it has recently achieved remarkable results~\cite{xie2019self} on ImageNet~\cite{deng2009imagenet}.  Following the procedure in~\cite{xie2019self}, we first train a teacher object detector $f(\mathbf{I},\theta_t^*)$ from the labeled images, where $\theta_t^*$ stands for the network weights. We then produce pseudo boxes for each unlabeled Web image $\widetilde{\mathbf{I}}_i\in\mathcal{U}$:
\begin{align}
\widetilde{\mathcal{T}_i}, \widetilde{\mathcal{S}}_i \leftarrow f(\widetilde{\mathbf{I}}_i, \theta_t^*), \quad i=1,...,m
\end{align}
where each pseudo box $\widetilde{t}_i^j \in \widetilde{\mathcal{T}}_i$ is also associated with a confidence score $\widetilde{s}_i^j\in \widetilde{\mathcal{S}}_i$. We obtain the confidence score from the detector's classification head. Finally, we train a student detector in a pre-training-and-fine-tuning manner. The idea is to pre-train the student detector using the Web images $\mathcal{U}$ along with the pseudo bounding boxes, followed by fine-tuning it on the curated set $\mathcal{D}$.

Modern object detectors generate hundreds of object candidates per image to ensure high recall even after non-maximum suppression~\cite{ren2015faster}, implying that many of the predicted pseudo boxes are incorrect. Traditional self-training used in image classification~\cite{xie2019self,scudder1965probability} disregards low-confidence labels when they train the student model. In the same spirit, we only keep the pseudo boxes whose confidence scores are higher than 0.7.

\subsection{Selective self-training for object detection (S$^2$OD)} \label{sec:s2od}

\begin{figure*}
\vspace{-10pt}
  \centering
  \includegraphics[width = 0.98\textwidth]{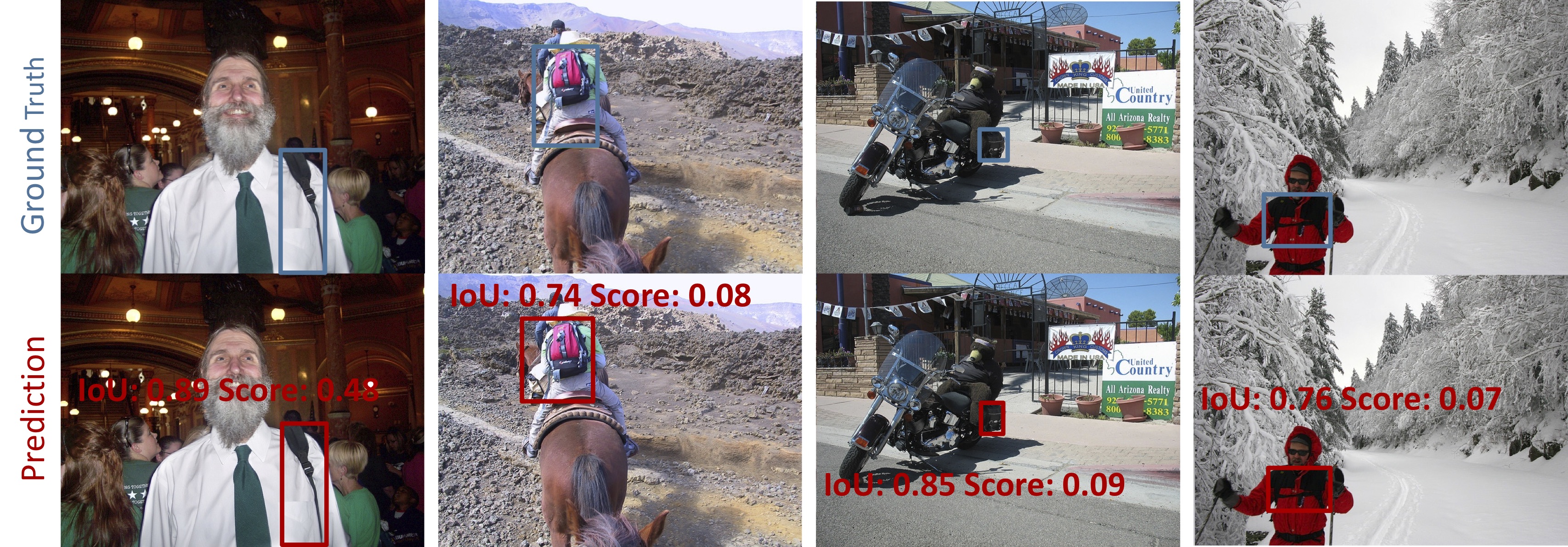}
  \vspace{-10pt}
  \caption{Top: images in the COCO training set with groundtruth boxes. Bottom: boxes, their IoUs with the groundtruth, and confidence scores predicted by the teacher detector. The correlation between IoU and confidence is low.}
  \label{fig:mismatch}
  \vspace{-10pt}
\end{figure*}

In SOD described above, a crucial step is the selection of pseudo boxes by thresholding the confidence scores. The effectiveness of SOD largely depends on the quality of the selected boxes, which, unfortunately, poorly correlates with the confidence score. Figure~\ref{fig:mismatch} shows some examples where the pseudo boxes tightly bound the backpacks, but the teacher detector assigns them very low confidence scores. As a result, those boxes would be removed before SOD trains the student detector, under-utilizing the Web images $\mathcal{U}$. What's worse is that the mistakenly removed boxes could be discovered as false hard negatives during
training. 

\eat{
Next, one image may have multiple boxes, and filtering some true positive boxes with low confidences may confuse the detector since the detector have to treat those boxes as background. Moreover, Vanilla-SD is sensitive to the threshold selection in (see the analysis in Section~\ref{sec:comparision}).
{There are several defects that can easily compromise Vanilla-SD. First, the confidence score is not closely correlated with the precision of selected boxes, meaning that confidence score is not a good indicator for box selection. Second, one object could have multiple predicted boxes, boxes with low confidence score are filtered out and thus treated as negative background during training(We need to show some example figures). Moreover, threshold on confidence score is very sensitive hyperparameter and needs to be tuned carefully(See the analysis in Section~\ref{sec:comparision}). This sensitivity means that we need to find a per class threshold(hyperparameter), which is impossible to optimize}}

\eat{Then it comes to the questions: Can we train a classifier to decide if the bounding box predicted by teacher model is a good candidate? And can we find a non-sensitive, class-agnostic parameter to classify these boxes?}

To tackle the challenges, we propose a selective net $g$ to calibrate the confidence scores of the pseudo boxes in the Web images $\mathcal{U}$. The main idea is to automatically group the  boxes into three categories: positive,  negative, and ambiguity. Denote by $g\circ \widetilde{\mathcal{T}_i}$ the grouping results of the pseudo boxes in $\widetilde{\mathcal{T}_i}$ for the $i$-th Web image $\widetilde{\mathbf{I}}_i$. We pre-train the student detector $f(\mathbf{I},\theta_s^*)$ by the following (and then fine-tune it on the curated training set $\mathcal{D}$),
\begin{align}\label{eq:sesd}
\theta_s^* \leftarrow \argmin\limits_{\theta_s} \; \frac{1}{m}\sum_{i=1}^m \ell(g \circ \widetilde{\mathcal{T}}_i,f(\widetilde{\mathbf{I}}_i,\theta_s)),
\end{align}
where $\ell(\cdot,\cdot)$ is the conventional loss for training object detectors. For Faster-RCNN~\cite{ren2015faster}, the loss consists of regression, classification, objectiveness, etc. All the positive boxes predicted by our selective net $g$ are used to activate those loss terms. In contrast, the ambiguous boxes, which could be correct but missed by the selective net, create a safe zone and do not contribute to any of the loss terms. This safe zone is especially useful when the learning algorithm has a hard negative mining scheme built in because it excludes potentially false ``hard negatives'' that fall in this ambiguity group. 

\vspace{-5pt}
\subsubsection{Preparing training data for the selective net $g$.} How do we learn the selective net $g$ without knowing any groundtruth labels of the Web images? We seek answers by revisiting the labeled training set $\mathcal{D}$ instead. After we learn the teacher detector, we apply it to the training images in $\mathcal{D}$ and obtain a large pool of pseudo boxes. We assign each pseudo box $\widetilde{t}_i^j$ in the $i$-th image to one of the three groups by comparing it to the groundtruth boxes $\mathcal{T}_i$, 
\begin{align}\label{eq:IoU_class}
g\circ \widetilde{t}_i^j = 
\left\{
\begin{aligned}
\text{Negative} & , & \max_{t\in  \mathcal{T}_i} \,IoU(\widetilde{t}_i^j, t)) \leq \gamma_l, \\
\text{Positive} & , & \max_{t\in\mathcal{T}_i}\,IoU(\widetilde{t}_i^j, t)) \geq \gamma_h, \\
\text{Ambiguity} & , & \text{otherwise},
\end{aligned}
\right.
\end{align}
where $IoU$ is the intersection-over-union, a common evaluation metric in object detection and semantic segmentation~\cite{lin2014microsoft,everingham2010pascal}, and $\gamma_h=0.6$ and $\gamma_l=0.05$ are two thresholds for IoU as opposed for the confidence scores. Interestingly, we can choose $\gamma_h$ by using the COCO evaluation protocol~\cite{lin2014microsoft} as follows. Considering all the boxes in the positive group as the teacher detector's final output, we can compute their mean average precision (mAP) over the labeled images $\mathcal{D}$. We choose the threshold of $\gamma_h=0.6$ that maximizes the mAP.

\vspace{-5pt}
\subsubsection{Preparing features for the selective net $g$.} 
We accumulate all potentially useful features to represent a pseudo box so that the selective net can have enough information to group the boxes. The feature vector for a box $\widetilde{t}_i^j$ is
\begin{align} \label{eq:features}
\left(f_{RoI}(\widetilde{t}_i^j,\mathbf{I}_i,\theta_t^*), \widetilde{s}_i^j, \widetilde{x}_i^j/W_i,\widetilde{y}_i^j/H_i,\widetilde{w}_i^j/W_i,\widetilde{h}_i^j/H_i, W_i, H_i\right)
\end{align}
where $f_{RoI}(\widetilde{t}_i^j,\mathbf{I}_i,\theta_t^*)$ is the RoI-pooled features~\cite{ren2015faster} from the teacher detector, $\widetilde{s}_i^j$ is the confidence score, $W_i$ and $H_i$ are respectively the width and height of the $i$-th image, and the others are normalized box coordinate and size. 

\vspace{-5pt}
\subsubsection{Training the selective net $g$.}
With the training data (eq.~(\ref{eq:IoU_class})) and features (eq.~(\ref{eq:features})) of the pseudo boxes, we learn the selective net by a three-way cross-entropy loss. We employ a straightforward architecture for the selective net. It comprises two towers. One is to process the normalized RoI-pooled features, and the other is to encode the remaining box features. They are both one-layer perceptrons with 512 and 128 hidden units, respectively. We then concatenate and feed their outputs into a three-way classifier.

One may concern that applying the teacher detector to the original training set $\mathcal{D}$ may not give rise to informative training data for the selective net because the detector could have ``overfitted'' the training set. Somehow surprisingly, we find that it is extremely difficult to overfit a detector to the training set, probably due to inconsistent human annotations of the bounding boxes. At best, the detector plays the role of an ``average rater'' who still cannot reach 100\% mAP on the training set whose bounding boxes are provided by different users. 

\subsection{Selective self-supervised self-training  object detection (S$^4$OD)} \label{sec:s4od}
Finally, we boost S$^2$OD by a self-supervised loss based on two considerations. One is that Xie et al.\ demonstrate that it is beneficial to enforce the student network to learn more knowledge than what the teacher provides (e.g., robustness to artificial noise)~\cite{xie2019self}. The other is that Jeong et al.\ show the effectiveness of adding a consistency regularization to semi-supervised object detection~\cite{jeong2019consistency}. 

More concretely, we add the following loss to eq.~(\ref{eq:sesd}) for each Web image $\widetilde{\mathbf{I}}_i$,
\begin{align}
    \ell_{i} :=  \sum_{\widetilde{t}_i^j \in \widetilde{\mathcal{T}}_i(\text{Positive}) } \left \| f_{RoI}(\widetilde{t}_i^j,\widetilde{\mathbf{I}}_i, \theta_s) - f_{RoI}(\vec{t}_i^j,\vec{\mathbf{I}_i},\theta_s)       \right \|_2 + \ell(g \circ \vec{\mathcal{T}}_i,f(\vec{\mathbf{I}}_i,\theta_s)), 
\end{align}
which consists of a consistency term borrowed from \cite{jeong2019consistency} and the same detection loss as eq.~(\ref{eq:sesd}) yet over a transformed Web image $\vec{\mathbf{I}}_i$ --- we explain the details below. They are additional cues to the pseudo boxes provided by the teacher. By learning harder than the teacher from all the supervision using the extra Web data, we  expect the student detector to outperform the teacher.

Given a noisy Web image $\widetilde{\mathbf{I}} \in \mathcal{U}$, we use the selective net $g$ to pick up  positive boxes $\widetilde{\mathcal{T}}(\text{Positive})$ and limit the consistency loss over them. This small change from~\cite{jeong2019consistency}, which is feasible due to the selective net, turns out vital to the final performance. We transform a Web image $\widetilde{\mathbf{I}}$ to $\vec{\mathbf{I}}$ by randomly choosing an operation from $\{\text{rotation by 90, 180, or 270 degrees}\}\times \{\text{horizontal flip or not}\}\times \{\text{random crop}\}$. We use the crop ratio 0.9 and always  avoid cutting through positive boxes. Accordingly, we can also obtain the transformed pseudo boxes $\vec{t}^j\in\vec{\mathcal{T}}$. The bottom panel of Figure~\ref{fig:overview} exemplifies this transformation procedure.

\section{Related Work}
\vspace{-6pt}


\subsubsection{Weakly supervised object detection.}

The great success of the state-of-the-art object detectors~\cite{zhou2019objects,girshick2015fast,ren2015faster,lin2017feature,liu2016ssd,redmon2017yolo9000,he2017mask} heavily relies on large volume of human annotated data.
For instance, the flagship COCO dataset~\cite{lin2014microsoft} contains about 10k boxes per class. 
Acquiring more curated data is challenging due to the high financial and time costs.

Accordingly, weakly supervised object detection (WSOD)~\cite{arun2019dissimilarity,bilen2016weakly,yang2019towards,tang2018weakly,tang2017multiple,zeng2019wsod2} --- learning to localize objects with image-level annotations only - has become an active research topic, since image-level annotations are easier to obtain than bounding box annotations.
Representative works~\cite{yang2019activity,prest2012learning,singh2019you,kumar2016track}
utilize motion cues in videos to delineate objects and refine object proposals.  In addition, Tao~\textit{et al.}~\cite{tao2018zero} incorporate web images to learn a good feature representation for WSOD. Fine-grained segmentation~\cite{gao2019c,li2019weakly} can also be used to guide WSOD.
Although great progress has been made in WSOD, there is still quite a gap for them to catch up its supervised counterpart.
The performance of fully supervised methods~\cite{wu2019detectron2,ren2015faster} are about 25 points better in terms of mean average precision compared to the weakly supervised ones~\cite{zeng2019wsod2}.

\vspace{-13pt}
\subsubsection{Semi-supervised Classification.}
The majority of data samples in the real-world lacks annotations. Hence, semi-supervised learning~\cite{bachman2014learning,rasmus2015semi,laine2016temporal,tarvainen2017mean,miyato2018virtual,luo2018smooth,berthelot2019mixmatch,xie2019unsupervised} exploits the potential of unlabeled data to gain more understanding of the population structure in general. 
Most of the works~\cite{luo2018smooth,berthelot2019mixmatch,xie2019unsupervised,miyato2018virtual} are based on consistency training, which constrains model predictions to be invariant to the noise injected to the input, hidden states, or model parameters. Consistency regularization~\cite{laine2016temporal,tarvainen2017mean,miyato2018virtual} has shown state-of-the-art performance in semi-supervised  classification~\cite{oliver2018realistic}. Besides, pseudo-label based approaches have improved the performance of semi-supervised learning by utilizing high-confident samples with pseudo-labels in training~\cite{iscen2019label,shi2018transductive,lee2013pseudo,arazo2019pseudo}. 

Xie \textit{et al.}~\cite{xie2019self} argue that consistency training in the early phase regularizes models towards high-entropy predictions and pseudo label based approaches rely on a model being trained rather than a high-accuracy converged model, which all prevent those methods from achieving better performance. They~\cite{xie2019self} instead utilize self-training~\cite{scudder1965probability,yalniz2019billion,riloff2003learning,yarowsky1995unsupervised} and aggressively inject noise to make the student better. They report state-of-the-art results on ImageNet with 300M unlabeled images. Hence, we build our approach upon self-training in this paper.

\vspace{-13pt}
\subsubsection{Semi-supervised Object Detection.}
The semi-supervised object detection approaches can be divided into two categories: \textbf{weakly semi-supervised detectors}~\cite{yan2017weakly,tang2016large,gao2019note,yang2019detecting,ramanathan2020dlwl} and \textbf{complete semi-supervised detectors}~\cite{jeong2019consistency,wang2018towards}. Weakly semi-supervised object detection method uses fully annotated data with box-level annotations as well as weakly labeled data with only image-level annotations. Tang~\textit{et al.}~\cite{tang2016large} propose an LSDA-based method that can handle disjoint sets in semi-supervised detection. Note-RCNN~\cite{gao2019note} proposes a mining and training scheme using a few seedbox-level annotations and a large scale of image-level annotations. Recently, Yang~\textit{et al.}~\cite{yang2019detecting} propose a fine-grained detection method that requires only bounding box annotations of a smaller number of coarse-grained classes and image-level labels on a large number of fine-grained classes.

Compared with weakly semi-supervised detectors, the complete semi-supervised detectors are more general by using unlabeled data in combination with the box-level labeled data. Our approach can be technically categorized into complete semi-supervised detectors. There are only a few research works in the complete semi-supervised detection field. Wang~\textit{et al.}~\cite{wang2018towards} present a principled Self-supervised Sample Mining (SSM) process in active learning to get reliable region proposals for enhancing the object detector. But they need additional human labeling effort to annotate the low-consistency samples. Jeong~\textit{et al.}~\cite{jeong2019consistency} introduce a consistency loss based method in semi-supervised object detection. They propose to add a simple consistency loss between original box and flipped box for the classification network. It is consistently effective for one-stage detectors but have limited performance improvement for two-stage detectors as box selection is not incorporated for unlabeled data. This pioneer work inspires us to dig deeper into two-stage detectors to build a more robust learning system with components like box selection.  Besides, we use the crawled Web images as the unlabeled data where the data distribution is unknown, while ~\cite{wang2018towards,jeong2019consistency} use COCO~\cite{lin2014microsoft} and PASCAL VOC 2012~\cite{everingham2010pascal} as unlabeled data where the data/class distributions is similar to their labeled dataset -- PASCAL VOC 2007~\cite{everingham2010pascal}.

\subsubsection{Unsupervised/Self-supervised Representation Learning.}
\vspace{-13pt}
Unsupervised/self-supervised representation~\cite{doersch2015unsupervised,he2019momentum,gan2016you,gan2016webly,gan2019self,zhao2018sound} learning on unlabeled data has attracted a great deal of attention nowadays. 
Some of them define a wide range of pretext tasks like recovering the input under some corruption~\cite{vincent2008extracting,pathak2016context}, predicting rotation~\cite{gidaris2018unsupervised} or patch orderings~\cite{doersch2015unsupervised,noroozi2016unsupervised} of an exemplar image, and tracking~\cite{wang2015unsupervised} or segmenting objects~\cite{pathak2017learning} in videos. Others utilize contrastive learning~\cite{oord2018representation,donahue2019large,he2019momentum,wu2018unsupervised,chen2020simple} by maximizing agreement between differently augmented views of the same data example. Good visual representations can help object detection~\cite{mahajan2018exploring,gidaris2018unsupervised}, and self-supervised learning has been applied to replace the supervised ImageNet pretraining~\cite{pathak2017learning,jenni2018self} for object detection. In addition, Lee~\textit{et al.}~\cite{lee2019multi} propose a set of auxiliary tasks to make better use of given limited labels. However, \cite{lee2019multi} requires box-level annotations to serve auxiliary task learning.


\section{Experiments}
We augment the COCO-backpack and COCO-chair training images with unlabeled Web images and run extensive experiments to test our approach on them. We note that the Web images are out of the distribution of COCO~\cite{lin2014microsoft}. Some of them may contain no backpack or chair at all, and we rely on our selective net to identify the useful pseudo boxes in them produced
by a teacher detector. 

Besides, since S$^4$OD is readily applicable to vanilla semi-supervised object detection, whose labeled and unlabeled sets follow the same distribution, we also test it following the experiment protocol of~\cite{jeong2019consistency}.

\subsection{Augmenting COCO-backpack(chair) with Web-backpack(chair)}\label{sec:comparision}

\begin{table}
\centering
\caption{Comparison results for detecting backpacks and chairs.}
\label{tab:comparision}
\vspace{-10pt}
\begin{tabular}{l|l|c|c|c|c|c|c|c}
\hline
 Dataset & Method & Web & $AP@[.5,.95]$  & $AP@.5$  & $AP@.75$ & $AP_{S}$  & $AP_{M}$ & $AP_{L}$  \\
\hline
\multirow{7}{*}{\shortstack{COCO-\\backpack}} 
 & R101-FPN~\cite{lin2017feature} & \xmark& 16.45 &33.63 &15.26 &18.29 &18.23 &17.61 \\
 
 & BD~\cite{lin2017feature} & \xmark & 17.34 & 34.71 & 15.13 & 19.03 & 19.46 & 21.31 \\

 & SOD~\cite{scudder1965probability} & \cmark &17.62 &33.83 &16.71 &18.98 &21.88 & 15.93 \\
 
& CSD~\cite{jeong2019consistency} & \cmark &17.87 &32.97 &16.55 & 20.42 & 22.05 & 20.45 \\
\cline{2-9}
& CSD-Selective (ours) & \cmark & 18.32 & 33.35 & 17.05 & \textbf{20.84} & 21.08 & 21.11 \\

 & S$^2$OD (ours) & \cmark & 18.28  & 35.35 & 18.46 & 19.23 & 21.76 & \textbf{23.94}  \\

 &S$^4$OD (ours) & \cmark & \textbf{19.48} &\textbf{36.25} & \textbf{19.01} & {20.31} & \textbf{23.47} & 18.53\\

\hline
\hline
\multirow{7}{*}{\shortstack{COCO-\\chair}} 
 & R101-FPN~\cite{lin2017feature} & \xmark &28.28 &48.93 &28.57 &19.14 &33.21 &42.06 \\
 
  & BD~\cite{lin2017feature}  & \xmark &  29.56 & 49.15 &30.70 &20.16 &36.39 &41.99 \\

  & SOD~\cite{scudder1965probability} & \cmark &30.01 & 49.05 & 31.53 & 20.32 & 37.10 & 43.12 \\
  
& CSD~\cite{jeong2019consistency} & \cmark &30.19  &48.58  &31.92  &20.77  &37.19 &43.74 \\
\cline{2-9}

& CSD-Selective (ours) & \cmark & 30.95 & 49.75 &32.11 &21.03 &38.02 &44.96\\

  &S$^2$OD (ours) & \cmark &30.95 & 50.25 & 31.92 & \textbf{21.19} & 37.82 & 45.48  \\
  &S$^4$OD (ours) & \cmark &\textbf{31.74} &\textbf{51.15} &\textbf{33.29} &20.57 &\textbf{38.66} & \textbf{46.93}\\
\hline
\end{tabular}
\vspace{-10pt}
\end{table}

We compare S$^4$OD to the following competing methods.
\begin{description}
\item[R101-FPN~\cite{lin2017feature}:] We use R101-FPN implemented in Detectron2~\cite{wu2019detectron2} as our object detector. Its $AP$ on COCO-validation is 42.03, which is on  par with the state of the arts. However, this detector still performs unsatisfactorily on the backpack and chair classes. As shown in Table~\ref{tab:comparision}, it only achieves 16.45 $AP$ on detecting backpacks and 28.28 $AP$ on chairs.
\item[BD~\cite{lin2017feature}:] We finetune R101-FPN on COCO-backpack and COCO-chair, respectively, by changing the 80-class detector to a binary backpack or chair detector. We observe that the binary detector outperforms its 80-class counterpart by about 1\% (0.9\% $AP$ for backpacks and 1.3\% $AP$ for chairs). We denote by BD the binary backpack (chair) detector and use it as the teacher detector for the remaining experiments.
\item[CSD~\cite{jeong2019consistency}:] We include the recently published consistency-based semi-supervised object detection (CSD) in the experiments. We carefully re-implement it by using the same backbone detector (R101-FPN) as ours. CSD imposes the consistency loss over all candidate boxes.
\item[CSD-Selective:] We also report the results of applying our selective net to CSD. We remove the loss terms in CSD over the boxes of negative and ambiguity groups predicted by the selective net. Table~\ref{tab:comparision} shows that it increases the performance of the original CSD by 0.45 $AP$ on detecting backpacks and 0.76 $AP$ on chairs. 
\item[SOD~\cite{scudder1965probability}:]  We presented a vanilla self-training procedure for object detection (SOD) in Section~\ref{sec:sod}. Unlike self-training in image classification~\cite{xie2019self}, we find that SOD is sensitive to the threshold of confidence scores probably because the out-of-distribution Web images make the scores highly uncalibrated. We test all thresholds between 0.5 and 0.9 (with an interval of 0.2) and find that SOD can only beat BD at the threshold of 0.7 (reported in  Table~\ref{tab:comparision}). 
\item[S$^2$OD (ours):] This is an improved method over SOD by employing the selective net (cf.\ Section~\ref{sec:s2od}). It is also an ablated version of our S$^4$OD by removing the self-supervised loss.
\end{description}

\begin{figure*}
\vspace{-10pt}
  \centering
  \includegraphics[width = 0.98\textwidth]{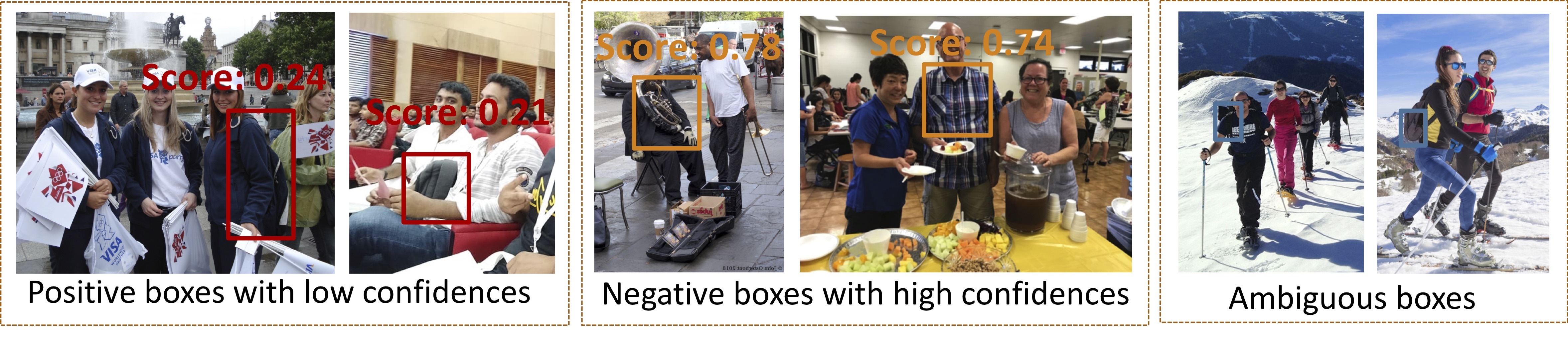}
  \vspace{-10pt}
  \caption{The two panels on the left: Positive (negative) boxes chosen by the selective net yet with low (high) confidence scores predicted by the teacher detector. Rightmost: Ambiguous boxes according to the selective net.}
  \label{fig:selectiveexample}
  \vspace{-10pt}
\end{figure*}

Table~\ref{tab:comparision} presents the comparison results on the validation sets of both COCO-backpack and COCO-chair. We can see that S$^2$OD performs consistently better than its teacher detector (BD), the vanilla self-training (SOD), and the consistency-based self-supervised learning (CSD). Our full approach (S$^4$OD) brings additional gains and gives rise to 19.48 $AP$ on detecting COCO backpacks and 31.74 $AP$ on detecting chairs --- about 2\% better than its original teacher (BD). Besides, the improvements of S$^2$OD over SOD and CSD-selective over CSD both attribute to the selective net. Finally, Figure~\ref{fig:selectiveexample} shows some cases where the selective net correctly groups low(high)-confidence  boxes into the positive (negative) group.

In addition to the overall comparison results in Table~\ref{tab:comparision}, we next ablate our approach and examine some key components. We also report the ``upper-bound'' results for augmenting COCO-backpack with Web-backpack by using the bounding box labels we collected for a subset of Web-backpack.

\begin{figure*}
\vspace{-10pt}
  \centering
  \includegraphics[width = 0.98\textwidth]{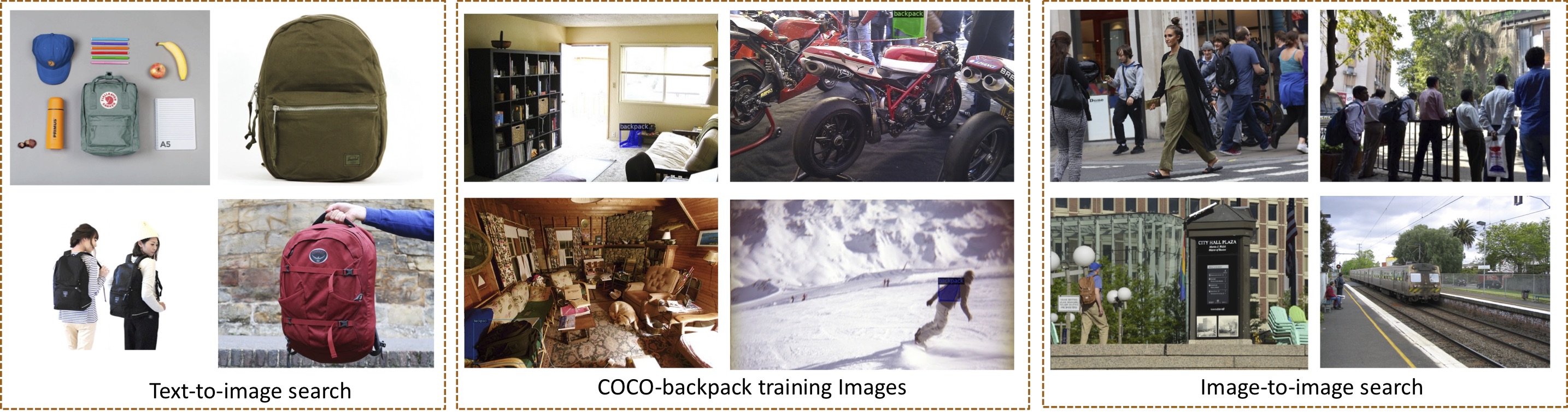}
  \vspace{-10pt}
  \caption{Example Web images for augmenting COCO-backpack.}
  \label{fig:textexample}
  \vspace{-25pt}
\end{figure*}

\vspace{-5pt}
\subsubsection{Web-backpack \textit{vs.}\ text-to-image search.}
The image-to-image search for Web images is the very first step of our approach, and it is superior over the text-to-image search in various aspects. As the leftmost panel in Figure~\ref{fig:textexample} shows, most top-ranked Web images are iconic with salient objects sitting in clean backgrounds if we search using class names. For COCO~\cite{lin2014microsoft}, the images are collected from Flickr by complex object-object and object-scene queries. Using the same technique, we can retrieve more natural images. However, they are mostly recent and come from diverse sources, exhibiting a clear domain shift from the COCO dataset which is about six years old. In contrast, the image-to-image search well balances between the number of the retrieved Web images and their domain similarity to the query images (cf.\ the right panel in Figure~\ref{fig:textexample}).

\begin{table}
\centering
\caption{Results of S$^4$OD with various Web data augmentations to COCO-backpack.}
\label{tab:ablation}
\vspace{-8pt}
\begin{tabular}{l | c| c|c| c|c| c}
\hline
 Method & $AP@[.5,.95]$  & $AP@.5$  & $AP@.75$ & $AP_{S}$  & $AP_{M}$ & $AP_{L}$  \\
\hline
 BD~\cite{lin2017feature}  & 17.34 & 34.71 & 15.13 & 19.03 & 19.46 & 21.31 \\
 S$^4$OD-text  &17.37 & 34.26 & 14.46 & 18.77 & 20.10 & 17.66  \\
 S$^4$OD w/ {1/3 Web-backpack} &18.40 & 34.12 & 17.91 & 19.25 & 21.20 & 21.56  \\
 S$^4$OD w/ {2/3 Web-backpack} & 19.08 &35.35 &17.96 &21.04 &20.15 &21.27  \\
 S$^4$OD w/ full Web-backpack & 19.48 &36.25 & 19.01 & 20.31 &
 \textbf{23.47} & 18.53\\
 S$^4$OD-$2_{nd}$ Iteration& \textbf{19.52} & \textbf{36.44}  & \textbf{19.07} & \textbf{21.70} & 20.90  & \textbf{24.90}  \\

\hline

\end{tabular}
\vspace{-10pt}
\end{table}

Table~\ref{tab:ablation} compares image-to-image search with text-to-image search by their effects on the final results. The S$^4$OD-text row is the results obtained using 230k Web images crawled by text-to-image. While they are slightly better than BD's results, they are significantly worse than what S$^4$OD achieves with  70k Web images retrieved by image-to-image search (cf.\ Table~\ref{tab:stat}).

\vspace{-5pt}
\subsubsection{Web-backpack: the size matters.}
We study how the number of the unlabeled Web images influences the proposed S$^4$OD by training with $1/3$ and $2/3$ of the crawled Web-backpack. As shown in table~\ref{tab:ablation}, S$^4$OD-1/3 can improve over BD by 1\% $AP$, and S$^4$OD-2/3 is better than BD by 1.7\% $AP$. In contrast, S$^4$OD with the full Web-backpack leads to about $2.1\%$ $AP$ improvement. Overall, we see that the more Web images, the larger boost in performance, implying that it is worth studying the data pipeline and expanding its coverage in future work.


\vspace{-5pt}
\subsubsection{Iterating S$^4$OD.} What happens if we use the detector trained by S$^4$OD as the teacher and train another student detector using S$^4$OD again? S$^4$OD-$2_{nd}$ Iteration in table~\ref{tab:ablation} outperforms the single-iteration version only by a very small margin. It is probably because, in the second iteration, the student detector does not receive more supervision than what the teacher BD provides. We will explore some stochastic self-supervised losses in future work to enforce the student to learn more than what the teacher provides at every iteration.


\begin{table}
\vspace{-10pt}
\centering
\caption{Comparison results on the \textbf{relabeled} COCO-backpack.}
\label{tab:comparerelabel}
\vspace{-10pt}
\begin{tabular}{l | c| c|c| c|c| c}
\hline
 Method & $AP@[.5,.95]$  & $AP@.5$  & $AP@.75$ & $AP_{S}$  & $AP_{M}$ & $AP_{L}$  \\
\hline

 BD  &18.75  & 36.03 & 16.98 & 11.90 & 21.62 & 31.58  \\
 Upper bound  &22.63 & 40.89 & 21.15 & 15.97 & 26.53 & 32.11\\
 S$^4$OD & 20.27 & 37.55 &20.49 & 13.51 & 24.04 & 31.17  \\
\hline
\end{tabular}
\vspace{-20pt}
\end{table}

\subsubsection{An ``upper bound'' of  S$^4$OD.} 
We further investigate the effectiveness of S$^4$OD by comparing it to an ``upper bound''. We run the experiments using the labeled subset of Web-backpack and the relabeled COCO-backpack to have consistent annotations across the two datasets. Recall that we have fixed some inconsistent bounding boxes in the original COCO-backpack in the relabeling process. As a result, if we  train BD~\cite{ren2015faster} on the relabeled COCO-backpack  and evaluate on the relabeled validation set, the $AP$ is 18.75 (cf.\ Table~\ref{tab:comparerelabel}), in contrast to 17.34 $AP$ (cf.\ Table~\ref{tab:comparision}) by BD trained and evaluated using the original COCO-backpack. Using this BD as the teacher, we train a student detector using S$^4$OD. By the ``upper bound'', we pool the labeled subset of Web-backpack and the relabeled COCO-backpack together and then train BD. The results in Table~\ref{tab:comparerelabel} indicate that S$^4$OD is almost right in the middle of the lower-bound (BD) and the upper bound. The gap between S$^4$OD and the upper bound is small. We will study better learning methods to close the gap and how the Web data volume could impact the performance in future work.

\subsection{Semi-supervised object detection on PASCAL VOC}
It is straightforward to extend S$^4$OD to multiple object detection tasks. Following the experiment protocol in~\cite{jeong2019consistency}, we further validate it in the setting of semi-supervised object detection, whose labeled and unlabeled sets are drawn from the same distribution. We use PASCAL VOC2007 trainval (5,011 images) as the labeled set and PASCAL VOC2012 trainval (11,540 images) as the unlabeled set. There are 20 classes of objects to detect. We test all detectors on the test set of PASCAL VOC2007 (4,952 images). In order to make our results comparable to what are reported in~\cite{jeong2019consistency}, we switch to Faster-RCNN~\cite{ren2015faster} with ResNet-50~\cite{he2016deep} as the base detector.

\begin{table}
\vspace{-10pt}
\centering
\caption{Comparison esults on  VOC2007  (* the numbers reported in~\cite{jeong2019consistency}) }
\label{tab:comonvoc}
\vspace{-10pt}
\begin{tabular}{l|l|c|c|c}
\hline
Method &Labeled data& Unlabeled data & $AP$ & Gain \\
\hline
Supervised~\cite{dai2016r,ren2015faster}&VOC07& -- & *73.9/74.1  & --\\
Supervised~\cite{dai2016r,ren2015faster}&VOC07\&12& -- & *79.4/80.3  & --\\
\hline
CSD~\cite{jeong2019consistency}&VOC07& VOC12 & *74.7  &  *0.8 \\
SOD &VOC07& VOC12 & 74.8 & 0.7 \\
S$^2$OD&VOC07& VOC12 & 75.8  & 1.7  \\
S$^4$OD&VOC07& VOC12 & \textbf{76.4}  & \textbf{2.3}  \\
\hline
\end{tabular}
\vspace{-10pt}
\end{table}

Table~\ref{tab:comonvoc} shows the comparison results. We include both our results and those reported in~\cite{jenni2018self} and mark the latter by *. Considering the object detector trained on VOC2007 as a baseline and the one trained on both datasets with labels as an upper bound, our S$^4$OD is right in the middle, outperforming CSD by 1.5 $AP$. Although we propose the selective net mainly to handle noisy Web images, it is delightful that the resulting method also works well with the clean, in-domain VOC2012 images. It is probably because the consistency loss in CSD drives the detector toward high-entropy, inaccurate predictions (cf.\ more discussions in~\cite{xie2019self}). Our selective net avoids this caveat by supplying high-quality boxes to the consistency loss.



\section{Conclusion}
In this paper, we propose a novel approach to improving object detection with massive unlabeled Web images. We collect the Web images with image-to-image search, leading to smaller domain mismatch between the retrieved Web images and the curated dataset than text-to-image search does. Besides, we incorporate a principled selective net into self-training rather than threshold confidence scores as a simple heuristic for selecting bounding boxes. Moreover, we impose a self-supervised training loss over the high-quality boxes chosen by the selective net to make better use of the Web images. The improvement in detecting challenging objects is significant over the competing methods, Our approach works consistently well on not only the Web-augmented object detection but also the traditional semi-supervised object detection. 

%
%


\end{document}


\pagestyle{headings}
\mainmatter
\def\ECCVSubNumber{6837}  

\title{Supplementary Materials for \\Improving Object Detection with Selective Self-supervised Self-training} 

\titlerunning{ECCV-20 submission ID \ECCVSubNumber} 
\authorrunning{ECCV-20 submission ID \ECCVSubNumber} 
\author{Anonymous ECCV submission}
\institute{Paper ID \ECCVSubNumber}

\maketitle

\appendix
We provide  the following to support the main text:
\begin{description}
\item[Section~\ref{app-protocol}:] description of the data annotation protocol used in this work (cf.\ Section 2 in the main paper),
\item[Section~\ref{app-ablation}:] additional experimental results of the self-training for object detection  (SOD) and our selective self-supervised self-training (S$^4$OD) (cf.\ Section 5.1 in the main paper), and
\item[Section~\ref{app-qualitative}:] qualitative  results of the binary detector (BD) and S$^4$OD.
\end{description}

\section{Data annotation protocol}\label{app-protocol}

We ask the raters to draw a bounding box around each and every backpack shown in the images, and the boxes should be tight. We underscore two scenarios in the instruction sent to the raters. 
\begin{enumerate}
    \item Please be careful about the labeling because many of the backpacks are very small and hard to
find. We want all of them to be labeled.
    \item Pay special attention to the backpack straps. Label them even if only the straps are visible. 
\end{enumerate}
Additionally, we provide concrete examples as shown in Figure~\ref{fig:insturction} to help the raters understand the annotation instruction. Figure~\ref{fig:cocovscloud} shows the annotations we collected \textit{vs.} COCO annotations.

\newpage 
\begin{figure*}
  \centering
  \includegraphics[width = 0.99\textwidth]{./supp_figures/COCOvsCloud.pdf}
  \vspace{-10pt}
  \caption{Our annotations \textit{vs.} COCO annotations.}
  \label{fig:cocovscloud}
\end{figure*}

\newpage 
\begin{figure*}
  \centering
  \includegraphics[width = 0.99\textwidth]{./figures/instruction.pdf}
  \vspace{-10pt}
  \caption{Some examples we provide to the raters to help them understand our annotation instruction.}
  \label{fig:insturction}
\end{figure*}

\newpage 
\section{Additional experimental results of SOD and S$^4$OD}\label{app-ablation}

\subsection{Threshold Selection}
The self-training for  object detection (SOD) selects pseudo boxes over unlabeled images by thresholding their confidence scores. We show in Table~\ref{tab:ablation} the SOD results with different thresholds: 0.5, 0.7 and 0.9. We find that thresholding the confidence scores can barely improve object detection, and it is sensitive to the threshold. In contrast, we threshold the intersection over union (IoU) of the pseudo and the groundtruth boxes over the training images for learning the selective net in S$^4$OD. The groundtruth boxes lend our selective net  robustness to the thresholds $\gamma_h$ and $\gamma_l$ (cf.\ Table~\ref{tab:ablation} which shows S$^4$OD with different choices of thresholding the IoU --- S$^4$OD-0.5-0.1 means $\gamma_h=0.5$ and $\gamma_h=0.1$). Especially, we can set  $\gamma_h$ by maximizing the detector's performance on the training set (cf.\ Section 3.2 in the main paper). 

\begin{table}
\centering
\caption{Results of SOD with different confidence score thresholds and S$^4$OD with different IoU thresholds.}
\label{tab:ablation}
\vspace{-8pt}
\begin{tabular}{l | c| c| c|c| c|c| c}

\hline
 \multirow{2}{*}{Method}& \multirow{2}{*}{$AP@[.5,.95]$} & Gain  & \multirow{2}{*}{$AP@.5$}  & \multirow{2}{*}{$AP@.75$} & \multirow{2}{*}{$AP_{S}$} & \multirow{2}{*}{$AP_{M}$} & \multirow{2}{*}{$AP_{L}$}  \\
 &   & Over BD &  &  &   &  &  \\
\hline
 BD~\cite{lin2017feature}  & 17.34 & -- & 34.71 & 15.13 & 19.03 & 19.46 & 21.31 \\
 SOD-0.5~\cite{scudder1965probability}  & 17.21 &-0.13  & 32.66 & 15.81 & 18.03 &21.03 & 18.09   \\
 SOD-0.7~\cite{scudder1965probability}  &17.37 & 0.03  & 34.26 & 14.46 & 18.77 & 20.10 & 17.66  \\
 SOD-0.9~\cite{scudder1965probability}  &17.22 & -0.12 & 32.76 & 17.44 & 19.03 & 21.05 &19.97  \\
 S$^4$OD-0.5-0.1 & 19.41 &2.07 & 36.19 & 18.67  & 21.07 & 22.08 & 21.02  \\
 S$^4$OD-0.6-0.05 &19.48 &2.14 &36.25 & 19.01 & 20.31 &  
 \textbf{23.47} & 18.53\\
 S$^4$OD-$2_{nd}$ Iteration& \textbf{19.52} & \textbf{2.18}&  \textbf{36.44}  & \textbf{19.07} & \textbf{21.70} & 20.90  & \textbf{24.90}  \\
\hline
\end{tabular}

\end{table}

\subsection{``Upper bounds'' of S$^4$OD}
We provide a ``upper bound'' for S$^4$OD tighter than the one described in the main text. We train S$^4$OD in two stages: pre-train the student detector on the unlabeled images and then fine-tune it on the labeled set. To obtain a more comparable ``upper bound'', we also train the binary detector in the two-stage manner except that we reveal labels to the detector in the first stage.  Table~\ref{tab:comparerelabel} shows that the two-stage ``upper bound'' is tighter than the one we provide in the main text. In the main paper, all the results of detecting backpacks using S$^4$OD  are obtained using the full Web-backpack. In Table~\ref{tab:comparerelabel}, we additionally show the results of S$^4$OD pre-trained using the ``Web-backpack labeled'' only --- but we use the images only and no labels at all.


\begin{table}
\centering
\caption{Comparison results on the \textbf{relabeled} COCO-backpack.}
\label{tab:comparerelabel}
\vspace{-10pt}
\begin{tabular}{l | c| c|c| c|c| c}
\hline
 Method & $AP@[.5,.95]$  & $AP@.5$  & $AP@.75$ & $AP_{S}$  & $AP_{M}$ & $AP_{L}$  \\
\hline

 BD~\cite{lin2017feature}  &18.75  & 36.03 & 16.98 & 11.90 & 21.62 & 31.58  \\
 
 Upper bound   &22.63 & 40.89 & 21.15 & 15.97 & 26.53 & 32.11\\
 Upper bound (two-stage) &  22.10 & 40.35 & 22.58 & 16.23 & 25.56 & 32.70 \\
 S$^4$OD w/ full Web-backpack& 20.27 & 37.55 &20.49 & 13.51 & 24.04 & 31.17  \\
 S$^4$OD w/ subset Web-backpack &19.70 & 38.33 & 18.05 &14.19 & 23.51 & 29.32 \\
\hline
\end{tabular}
\end{table}

\subsection{More iterations on VOC}
We train the second iteration of S$^4$OD on VOC benchmark~\cite{everingham2010pascal}. Table~\ref{tab:comonvoc} shows that S$^4$OD-$2_{nd}$ Iteration can improve the single-iteration result by a small margin. This is similar to our observation on COCO-backpack (cf. \ S$^4$OD-$2_{nd}$ Iteration in Table~\ref{tab:ablation}). We conjecture that it is because the student detector can not obtain more supervision from the teacher in the second iteration.

\begin{table}
\centering
\caption{Comparison results on  VOC2007  (* the numbers reported in~\cite{jeong2019consistency})}
\label{tab:comonvoc}
\vspace{-10pt}
\begin{tabular}{l|l|c|c|c}
\hline
Method &Labeled data& Unlabeled data & $AP$ & Gain \\
\hline
Supervised~\cite{dai2016r,ren2015faster}&VOC07& -- & *73.9/74.1  & --\\
Supervised~\cite{dai2016r,ren2015faster}&VOC07\&12& -- & *79.4/80.3  & --\\
\hline
CSD~\cite{jeong2019consistency}&VOC07& VOC12 & *74.7  &  *0.8 \\
S$^4$OD&VOC07& VOC12 & 76.4 & 2.3  \\
S$^4$OD-$2_{nd}$ Iteration&VOC07& VOC12 & \textbf{76.5}  & \textbf{2.4}  \\
\hline
\end{tabular}
\end{table}

\section{Qualitative results}\label{app-qualitative}
We show some qualitative results of the binary detectors (BDs) and S$^4$OD. They demonstrate the cases that S$^4$OD improves over BD and when it does not. They also underscore the challenges in detecting the common, ``uninteresting'' man-made objects. We apply our most powerful models on the validation set of COCO-backpack and COCO-chair and calculate the per-image $AP$ for each image. We then rank the images according to the per-image $AP$ so that we know which images contain easy-to-detect objects and which images have hard-to-detect objects. Figures~\ref{fig:topbackpack_1}\&\ref{fig:topbackpack_2} and \ref{fig:topchair_1}\&\ref{fig:topchair_2} show the easy-to-detect backpacks and chairs, respectively. Figure~\ref{fig:badbackpack_1}\&\ref{fig:badbackpack_2} and \ref{fig:badchair_1}\&\ref{fig:badchair_2} show the hard backpacks and chairs, respectively. Figure~\ref{fig:noobject} shows some predicted boxes on the images which contain no backpack or chair. Overall, S$^4$OD can provide more accurate boxes than BD. Especially, BD  generates many false positive predictions. The hard-to-detect objects are often small, occluded, or labeled incorrectly.

\begin{figure*}
  \centering
  \includegraphics[width = 0.90\textwidth]{./supp_figures/top_backpack_1.pdf}
  \vspace{-10pt}
  \caption{Backpacks that are easy to detect. We show the predicted boxes by S$^4$OD on the left, the predictions by BD in the middle, and the ground-truth on the right.}
  \label{fig:topbackpack_1}
\end{figure*}

\begin{figure*}
  \centering
  \includegraphics[width = 0.90\textwidth]{./supp_figures/top_backpack_2.pdf}
  \vspace{-10pt}
  \caption{Backpacks that are easy to detect. We show the predicted boxes by S$^4$OD on the left, the predictions by BD in the middle, and the ground-truth on the right.}
  \label{fig:topbackpack_2}
\end{figure*}

\begin{figure*}
  \centering
  \includegraphics[width = 0.90\textwidth]{./supp_figures/top_chair_1.pdf}
  \vspace{-10pt}
  \caption{Chairs that are easy to detect. We show the predicted boxes by S$^4$OD on the left, the predictions by BD in the middle, and the ground-truth on the right.}
  \label{fig:topchair_1}
\end{figure*}

\begin{figure*}
  \centering
  \includegraphics[width = 0.90\textwidth]{./supp_figures/top_chair_2.pdf}
  \vspace{-10pt}
  \caption{Chairs that are easy to detect. We show the predicted boxes by S$^4$OD on the left, the predictions by BD in the middle, and the ground-truth on the right.}
  \label{fig:topchair_2}
\end{figure*}

\begin{figure*}
  \centering
  \includegraphics[width = 0.90\textwidth]{./supp_figures/bad_backpack_1.pdf}
  \vspace{-10pt}
  \caption{Backpacks that are hard to detect. We show the predicted boxes by S$^4$OD on the left, the predictions by BD in the middle, and the ground-truth on the right.}
  \label{fig:badbackpack_1}
\end{figure*}

\begin{figure*}
  \centering
  \includegraphics[width = 0.90\textwidth]{./supp_figures/bad_backpack_2.pdf}
  \vspace{-10pt}
  \caption{Backpacks that are hard to detect. We show the predicted boxes by S$^4$OD on the left, the predictions by BD in the middle, and the ground-truth on the right.}
  \label{fig:badbackpack_2}
\end{figure*}

\begin{figure*}
  \centering
  \includegraphics[width = 0.85\textwidth]{./supp_figures/bad_chair_1.pdf}
  \vspace{-10pt}
  \caption{Chairs that are hard to detect. We show the predicted boxes by S$^4$OD on the left, the predictions by BD in the middle, and the ground-truth on the right.}
  \label{fig:badchair_1}
\end{figure*}

\begin{figure*}
  \centering
  \includegraphics[width = 0.85\textwidth]{./supp_figures/bad_chair_2.pdf}
  \vspace{-10pt}
  \caption{Chairs that are hard to detect. We show the predicted boxes by S$^4$OD on the left, the predictions by BD in the middle, and the ground-truth on the right.}
  \label{fig:badchair_2}
\end{figure*}

\begin{figure*}
  \centering
  \includegraphics[width = 0.90\textwidth]{./supp_figures/no_object.pdf}
  \vspace{-10pt}
  \caption{False positives over the images that contain neither backpacks nor chairs. We show the predicted boxes by S$^4$OD on the left and the predictions by BD on the right. }
  \label{fig:noobject}
\end{figure*}

\clearpage
%
%
\bibliographystyle{splncs04}
\bibliography{egbib}